\documentclass[journal]{IEEEtran}
\usepackage{times}
\usepackage{epsfig}
\usepackage{graphicx}
\usepackage{amsmath}
\usepackage{amssymb}
\usepackage{subcaption}
\usepackage{dirtytalk}
\usepackage{color}

\usepackage{algorithm}
\usepackage{algpseudocode}
\usepackage[font=normal,labelfont=bf]{caption}
\usepackage{enumitem}
\usepackage{tensor}
\usepackage{hyperref}
\hypersetup{
    colorlinks=true,
    linkcolor=blue,
    filecolor=magenta,      
    urlcolor=cyan,
}

\DeclareMathOperator*{\argmax}{arg\,max}

\newcommand{\norm}[1]{\left\lVert#1\right\rVert}

\ifCLASSINFOpdf
\else
\fi
\begin{document}
%
\title{Search Tracker: Human-derived object tracking in-the-wild through large-scale search and
retrieval}
%
%
%


\author{Archith J. Bency\thanks{Archith J. Bency and B.S. Manjunath are with the Department of Electrical and Computer Engineering, University of California, Santa Barbara, CA, 93106 USA e-mail: (\{archith, manj\}@ece.ucsb.edu).},
        S. Karthikeyan\thanks{S. Karthikeyan, Carter De Leo and Santhoshkumar Sunderrajan were with the Department of Electrical and Computer Engineering, University of California, Santa Barbara, CA, 93106 USA.},
        Carter De Leo,
        Santhoshkumar Sunderrajan and B. S. Manjunath, ~\IEEEmembership{Fellow,~IEEE}}

\maketitle

\begin{abstract}
Humans use context and scene knowledge to easily localize moving objects in conditions of complex illumination changes, scene clutter and occlusions. In this paper, we present a method to leverage human knowledge in the form of annotated video libraries in a novel search and retrieval based setting to track objects in unseen video sequences. For every video sequence, a document that represents motion information is generated. Documents of the unseen video are queried against the library  at multiple scales to find videos with similar motion characteristics. This provides us with coarse localization of objects in the unseen video. We further adapt these retrieved object locations to the new video using an efficient warping scheme. The proposed method is validated on in-the-wild video surveillance datasets where we outperform state-of-the-art appearance-based trackers. We also introduce a new challenging dataset with complex object appearance changes.
\end{abstract}

\begin{IEEEkeywords}
Visual object tracking, Video search and retrieval, Data-driven methods
\end{IEEEkeywords}

%
\IEEEpeerreviewmaketitle

\section{Introduction}
%
%
%
%

\IEEEPARstart{O}{bject} tracking is a well-studied computer vision problem. Tracking algorithms (or trackers) should be robust to  large
 variations of lighting,  scene clutter, and handle occlusions while localizing an
object across frames. A number of algorithms
\cite{JepsonFleet, appearance01}  have approached the problem of tracking by modeling the appearance of objects as
they go through illumination, pose and occlusion changes in image sequences. Motion
models are also incorporated in these algorithms to provide a prior for object location in the
current frame, given the state of the tracker in previous frames. Recent state-of-the-art algorithms have been tested on real-world
datasets \cite{viper, PETS2010, PETS2009}.
These datasets are usually of good image quality and capture sufficient visual
information to distinguish between the object of interest and its surroundings.
While tracking objects in videos with low-quality imaging, these
methods have difficulty in learning robust appearance and motion models. As video infrastructures like surveillance networks have
been around for a decade, it is still important to be able to detect and track
objects in legacy low-resolution, low-quality videos.

\begin{figure}
\centering
\includegraphics[width=\columnwidth]{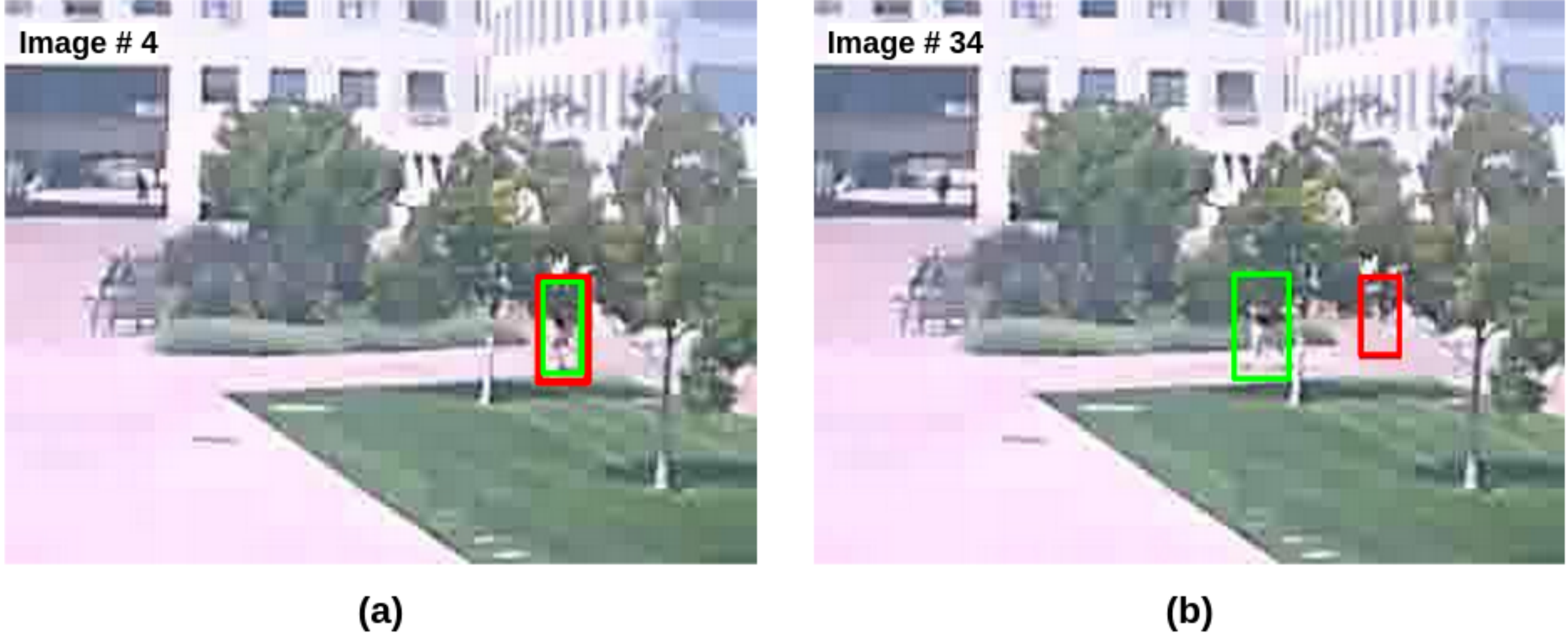}
\caption{Result frames for a sequence processed by VTD \cite{DBLP:conf/cvpr/KwonL10}, an appearance-based tracker, with a
pedestrian walking from right to left. Red and green
boxes represent the tracker's predicted object location and ground-truth
respectively. $\textit{Best viewed in color.}$ }
\label{fig:trackerFailure}
\end{figure}

An example of tracker failure, where appearance-based features are used, is presented
in Figure \ref{fig:trackerFailure}. The appearance-based
tracker gets distracted by background clutter of trees and learns an
incorrect appearance model. This leads to tracker failure and the object state
is lost. Also, most of the trackers
need either object detectors or manual initialization for the methods to start
tracking objects. Object detectors  \cite{DalalHog, Viola01rapidobject} are  prone to failure on low-quality images as detectors
trained on one dataset may not have good detection performance on a different
dataset. In conditions where one may come across a diverse set of objects (say
humans, vehicles, animals, etc.), a large number of detectors would be needed to
generate detections for the trackers to be effective.

Humans on the other hand, find tracking objects in such  scenarios
to be a relatively  easy task. Human-annotated bounding boxes are  of higher quality than those generated
by tracking algorithms. Humans leverage contextual knowledge of both the
scene and typical object motion to effortlessly track objects.
 Directly replicating human knowledge would involve coming up with complex
 computational models for tracking. This paper describes a method to leverage datasets of human annotated videos to
track moving objects in new videos, the Search Tracker (ST).
We maintain a library of training videos containing objects annotated with bounding boxes. The training
videos are then transformed into representative documents which are indexed along with the
provided bounding boxes. These documents encode motion patterns of annotated objects in the training videos.

For tracking to be applied on a new test video, we generate similar documents from this video. 
These documents are matched against the library documents to find video segments with similar motion patterns. The assumption is that video segments with similar motion characteristics will have similar object annotations. 
Finally,  object annotations corresponding to the retrieved results are transferred and warped to match the motion in the test video better.

The main contributions of this paper are:

\begin{itemize}[leftmargin=*]
    \item{We present a method that tackles the problem of tracking objects in-the-wild using a search and retrieval framework by learning long term motion patterns from a library of training videos.}
    \item{This approach carries out object tracking without dedicated object detectors or manual initialization and is automated in the true sense.}
    \item{This approach demonstrates an empirically effective way of transferring information learnt from one dataset to apply onto other datasets of very different visual contents such as view-points, types of objects, etc. }

\end{itemize}

 The rest of the paper is organized as follows. Section \ref{section:relatedWorks} presents an overview
 of related work. Section  \ref{section:approach} provides the details of the proposed
 method with a focus on the offline library generation and the online test video
 tracking process.  Section \ref{section:Experiments} elaborates on the experiments done to validate our approach and
  we present our comments, possible future work and conclusions in Section \ref{section:discConc} and Section \ref{section:futWork} respectively.

 \section{Related Work}\label{section:relatedWorks}
 Object tracking is an active research area in the computer vision community.
Surveys of object tracking algorithms are provided in \cite{ShahSurvey, LiSurvey, shahSurvey2}. A large number of tracking algorithms learn an appearance model from the initial frame and adapt it to information from incoming frames. Tracking results in the current frame are incorporated into the tracking model for subsequent frames. This online paradigm is called tracking-by-detection \cite{BMVC.20.6, saffari10a}. The simplest object trackers within this paradigm have used color histograms \cite{MeanShift} and template matching \cite{templateMatching}. However, these methods are susceptible to tracking errors which leads to the tracker model incorporating background clutter and occlusions. Multiple Instance Learners \cite{milboost} and trackers based on Structured-label SVMs \cite{struckTracker} have tackled the problem of sampling the right image patches for online learning. Yi et al. \cite{InitInsensitive} propose a visual tracker which is insensitive to the quality of manual initialization. The tracker takes advantage of motion priors for detected target features from optical flow, thereby handling inaccurate initializations. This method still relies on either a manual initialization or an object detector to initialize the tracker reliably in a close neighbourhood of the ground-truth to be successful.

Additionally there are methods that learn from annotated datasets in order to create priors which aid appearance based trackers. Manen et al. \cite{ManenECCV2014} have proposed an interesting framework which learns how objects typically move in a scene and uses that knowledge as a prior to guide appearance-based trackers to handle occlusions and scene clutter. This method requires annotations of multiple object tracks in the same scene. In contrast, our method can track objects in scenes totally unrelated to the dataset we learn from. Rodriguez et al. \cite{DataDrivenCrowdAnalytics} use a large database of crowd videos to search and find priors  in order to guide a linear Kalman filter based tracker. The method requires that the query video has similar scene appearance to retrieved library videos and that the target's position be manually initialized which are not required for the proposed approach.

On the front of biologically inspired systems, there are several works which leverage human contextual knowledge for computer vision tasks like action recognition ~\cite{Jhuang07abiologically} , scene classification ~\cite{journals/pami/SiagianI07, Song:2010:BIF:1821438.1821454}, and object detection ~\cite{papadopoulos2014training, karthikeyan2013and}.

\begin{figure}
\begin{center}
\includegraphics[width=\columnwidth]{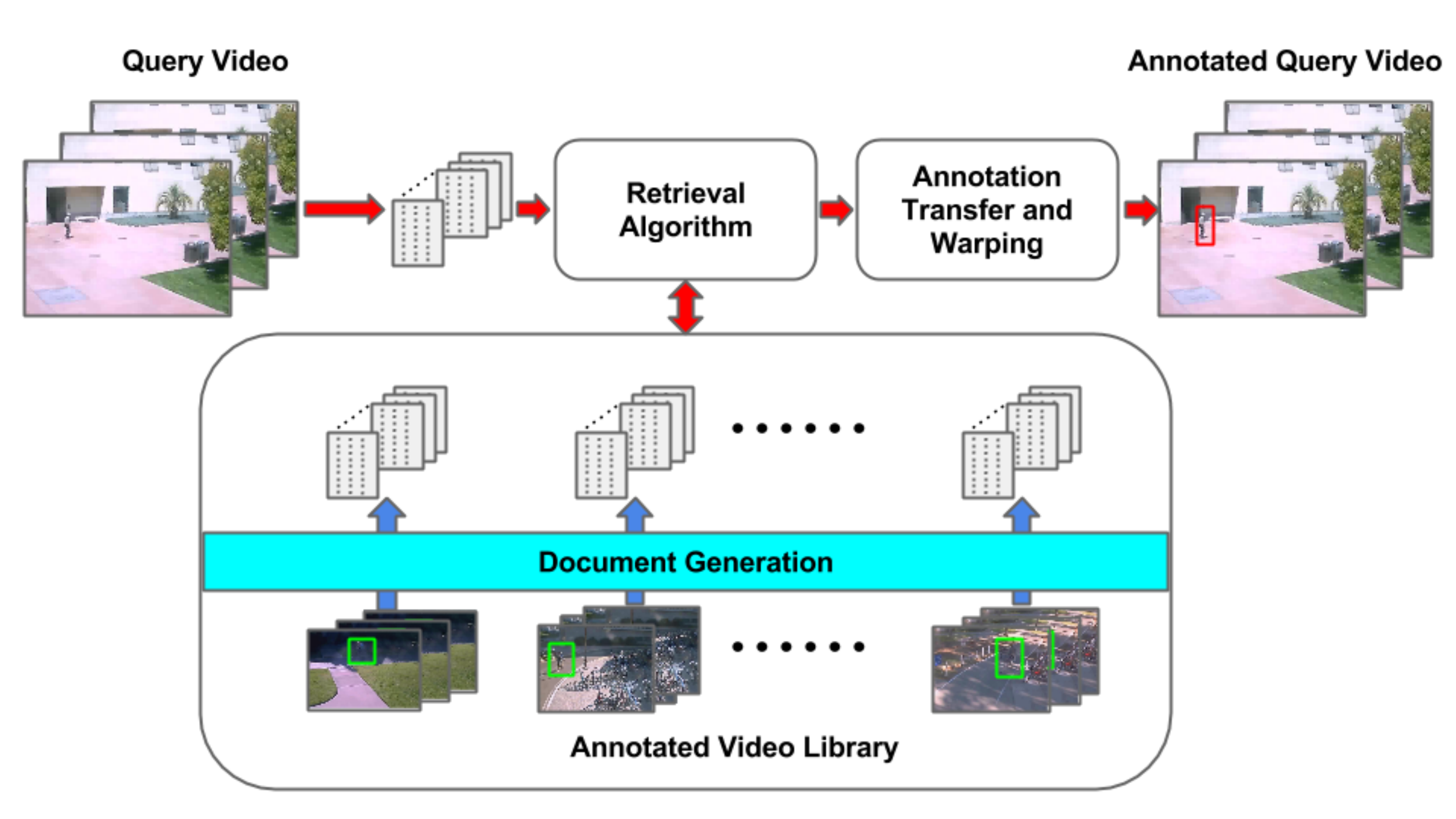}
\end{center}
\caption{Block diagram presenting a high-level view of the proposed system. Representative documents are generated from the query video. These documents encode object motion characteristics. Query documents are submitted to the retrieval algorithm to find matches. Annotations corresponding to found matches are then transferred and warped onto the query video.  Red arrows represent online steps and blue arrows represent off-line steps of the approach. $\textit{Best viewed in color.}$  }
\label{fig:BlockDiagram}
\end{figure}

\begin{figure}
\centering
\includegraphics[width=\columnwidth]{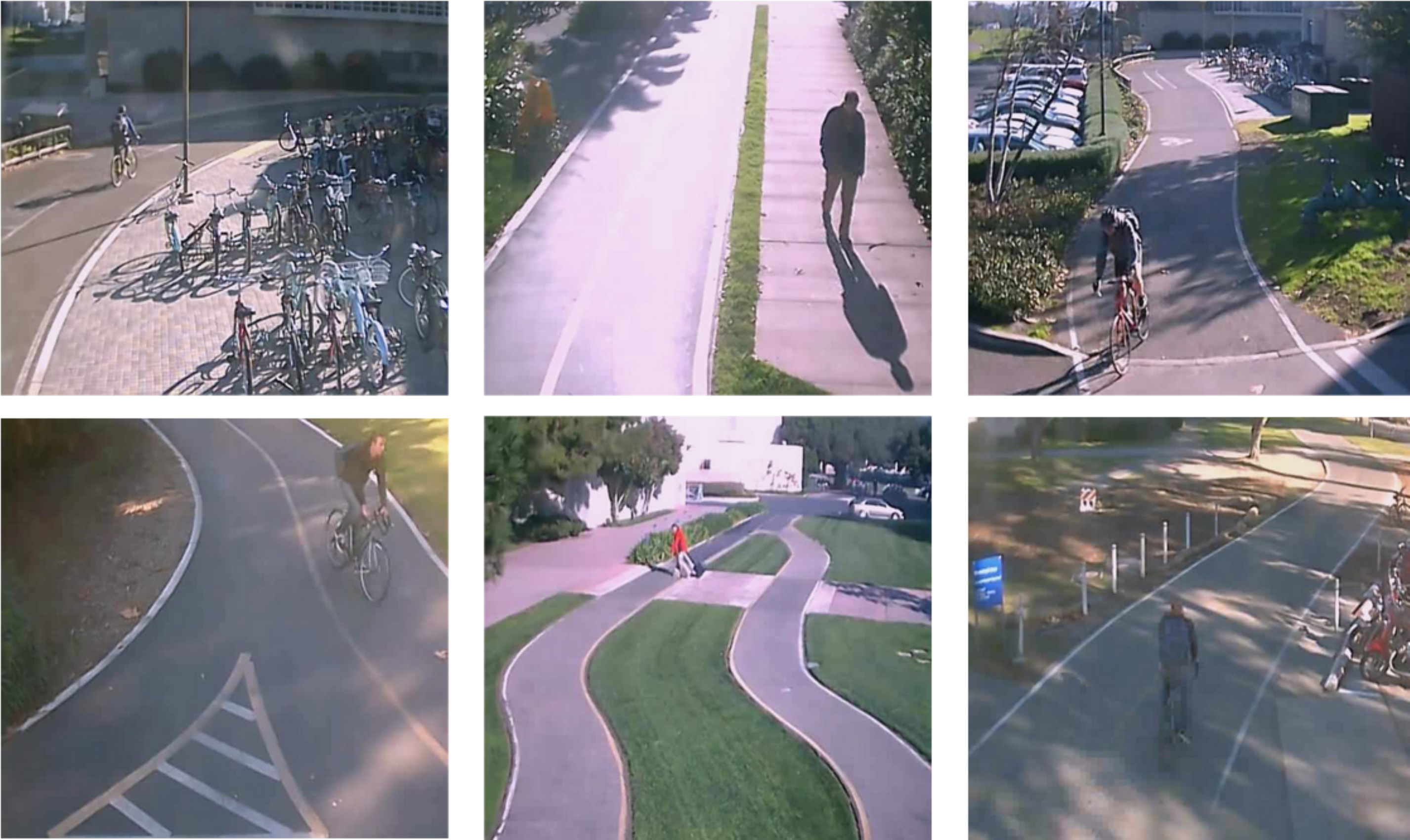}
\caption{Example frames from six of the videos included in the library of training videos. $\textit{Best viewed in color.}$ }
\label{fig:CamViews}
\end{figure}

\section{Search Tracker} \label{section:approach}
 
 \subsection{Overview of the Approach}
 
We aim to track objects in unseen videos by finding matches for motion patterns  amongst a library of videos with indexed human-generated annotations.
 There are two distinct phases in the proposed method. The offline phase operates on a library of training videos with annotated bounding boxes. Training videos are transformed into representative documents which are indexed along with the provided bounding boxes. The documents encode long-term motion patterns of annotated objects. We use optical flow \cite{SecretsofOF} to represent motion information from videos.

 During the second phase, a new test video is accepted for tracking. Documents similar to those created for the training videos are generated. These documents incorporate motion patterns across different scales and spatial locations, which can be matched to those in the training library. This enables the use of smaller training libraries to represent diverse motion patterns. The matching and retrieval process handles detection and tracking of multiple objects in the test videos.
 
Once matches for test video documents from the training
database are found, associated annotation bounding boxes
are transferred to the test video. Transferred bounding boxes are warped to improve the match with motion characteristics of tracked objects. We utilize non-maximal suppression to derive the best bounding boxes from the set of warped
bounding boxes. Subsequently, a smoothing step is carried out to regularize the scale of bounding boxes for the detected objects.

To summarize, human-generated annotations are leveraged to track moving objects in challenging scenarios without actual human review
of the test video. A high-level block diagram depicting the proposed
method is presented in Figure \ref{fig:BlockDiagram}. The library creation process and the proposed query scheme are explained below.

 \subsection{Offline library creation}
 \subsubsection {Training video library}
 The training video library consists of around 20 minutes of publicly available surveillance videos recorded across 10 camera views on the UCSB campus \cite{sunderrajan2013context, xu2013graph}. The resolution of the videos is 320$\times$240 and they are recorded at the rate of 24 frames per second. Note that this doesn't constrain the dimensions of test videos. The library videos capture scenes of pedestrians and bicyclists on campus bike-paths from various viewpoints. There are a total of 291 object tracks in the library. Example frames from the library are shown in Figure \ref{fig:CamViews}. Human-generated annotations corresponding to individual objects are stored and indexed. To increase the diversity of motion patterns in the dataset, we have generated horizontally and vertically flipped versions of library videos.

\begin{figure}
\centering
\includegraphics[width=0.8\columnwidth, height=8cm]{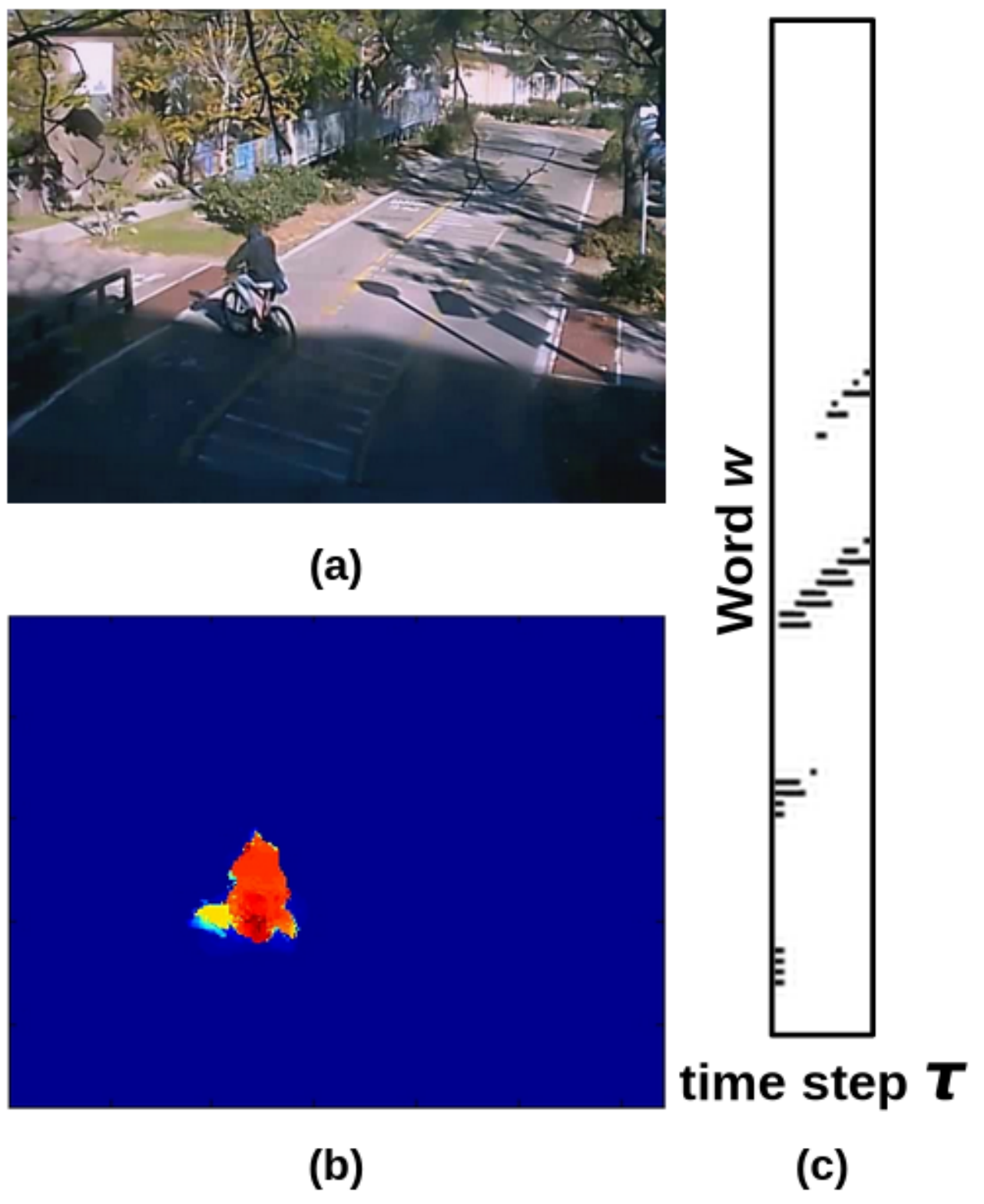}
\caption{(a) An example frame from a sequence belonging to the training video library. (b) Visualization of the optical flow  magnitude for the shown frame. (c) Document generated from the sequence. The vertical axis corresponds to the word, which in-turn corresponds to spatial location of a cube and the observed direction of motion. The horizontal axis corresponds to time steps. The document is binary valued with the black regions signifying activations. $\textit{Best viewed in color.}$  }
\label{fig:documentImg}
\end{figure}

 \subsubsection {Video document generation} \label{section:DocGen}
 
We divide the training videos into small non-overlapping spatio-temporal cubes and compute dense optical flow across frames \cite{SecretsofOF}. For each spatio-temporal cube, optical flow vectors are averaged over a time-step and those exceeding a specified magnitude  are binned into four directions (up, left, down and right). The binning is performed as a soft decision where an optical flow vector can belong to two directions (eg. left and up), the contribution being directly proportional to how close the vector is to these directions. The `votes' for each of the optical flow vectors are summed up and thresholded. This generates a 4-bit binary motion code for each cube. For our experiments, we have set the spatial size of cube to 20$\times$20 and the temporal step size to 4 frames. The spatial locations and the motion code of the cubes are flattened to a single column vector. Each of the binary codes in the column vector are termed as words with them being denoted by the variable $w \in [0, W)$. $W$ is the number of spatio-temporal cubes in a time step multiplied by the number of quantized directions. The value of $W$ is derived as:
\begin{equation}
W = \frac{I_{X}*I_{Y}*m}{c_{X}*c_{Y}}
\end{equation}
\noindent where $I_{X}$ and $I_{Y}$ are the video width and height, $c_{X}$ and $c_{Y}$ are the spatio-temporal cube width and height, and $m$ is the number of binary bits in the motion code. 
For our experiments, $W = 768$. We tried out different values for these design parameters and got the best performance for the values specified before. The horizontal axis represents time-steps in the video, indexed by $t \in [0, T)$.  An example document is shown in Figure \ref{fig:documentImg}. Design of the video document is meant to capture spatial location and directions of object motions from training videos.

 \subsubsection {Motion and track indexing}\label{subsec:IndexDesc}
 To enable search and retrieval of motion patterns from training videos, we divide the documents along the temporal dimension into fragments. We choose a parameter $T_f$ which denotes the document fragment length. This is the temporal duration of the basic retrievable segment of a library video that will be chosen and combined to represent a query video. A fragment is, hence, a contiguous subset of $T_f$ columns from a video document. In our experiments, we have fixed $T_f$ to 8. Each video fragment can be represented as a set of activated $(w, \tau)$ pairs, where $\tau \in [0, T_f)$ is the time relative to the start of the fragment. Each overlapping segment of a document with duration $T_f$ is indexed as an individual fragment. During training, the library data is stored and indexed across five database tables:

\begin{itemize}
  \item \textit{Fragment forward index}: This table contains a row for each fragment, mapping from a fragment name to its set of $(w, \tau)$ activations.
  \item \textit{Fragment inverse index}: This table contains a row for each $(w, \tau)$ pair, mapping to the fragment names in which that pair appears.
  \item \textit{Flow fields}: This table contains the optical flow magnitude for each time step in every document. These will be used later for warping.
  \item \textit{Track forward index}: This table contains a row for each unique track id present in the human-generated annotations, mapping to a bounding box for each frame where the corresponding object is present.
  \item \textit{Track inverse index}: This table contains a row for each fragment, mapping to the set of track ids annotated during that fragment's duration.
  
\end{itemize}

\begin{figure*}
\begin{center}
\includegraphics[width=2\columnwidth]{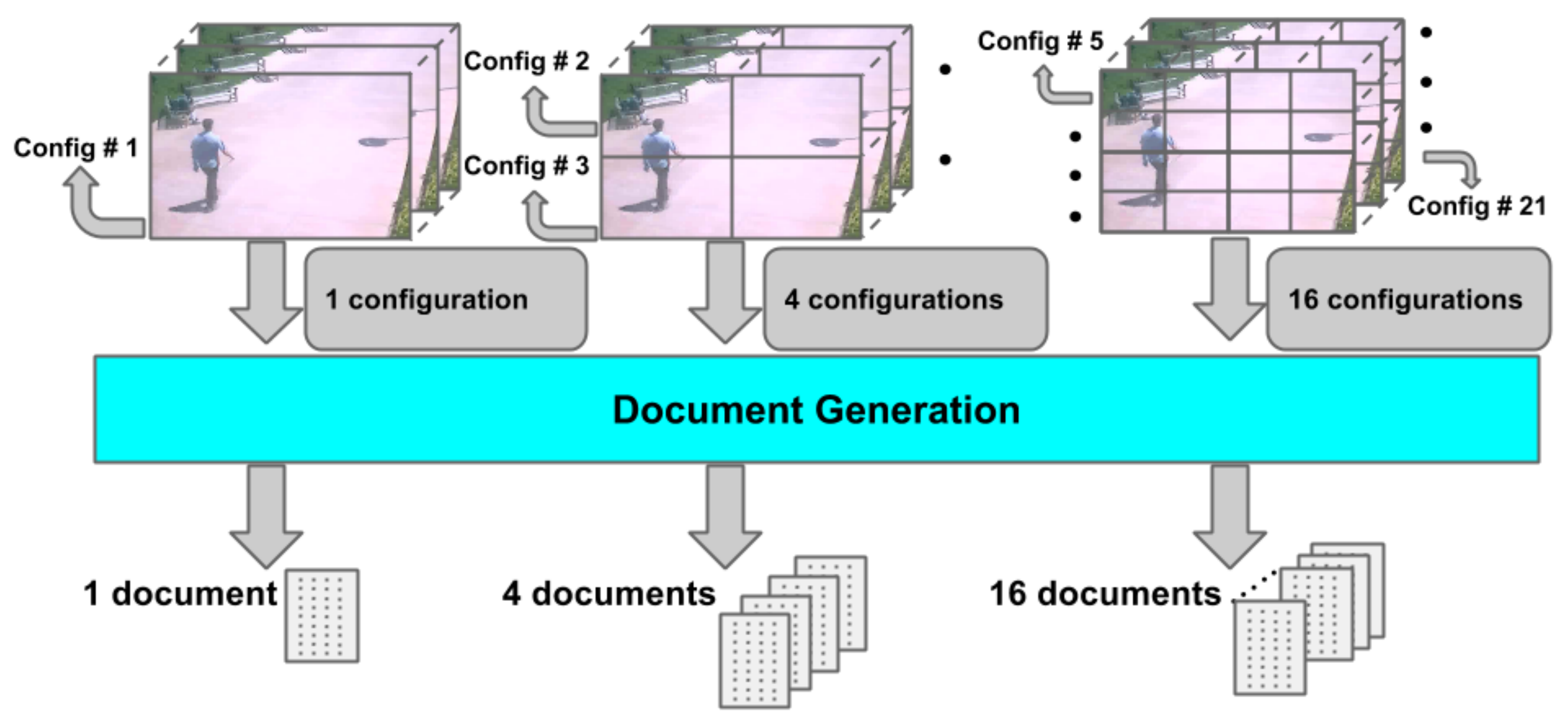}
\end{center}
\caption{The  query video gets processed as twenty one different spatio-temporal volume configurations as depicted above. Each configuration gets processed into a document individually. $\textit{Best viewed in color.}$}
\label{fig:onlineQuer}
\end{figure*}

\subsection{Online video queries}

With offline library creation steps complete, the system is ready to provide tracks for a new unseen input video. Keeping with the search and retrieval metaphor, an input video is called a query.

\subsubsection{Multi-scale video document generation}
 In order to be able to match motion patterns at multiple scales and spatial locations from the training video library, we generate documents for different configurations of the input video. The configurations are illustrated in Figure \ref{fig:onlineQuer}.
The first configuration has the video processed at the original scale. The next 4 configurations has the video spatially divided into 4 quadrants. The quadrants are individually processed to create one document each. Additional 16 configurations are generated by spatially dividing the video into 16 parts of identical sizes and each part generating a document. In total, for each video we generate 21 documents. The spatial dimensions of the spatio-temporal cubes used during document generation are modulated with size of the video configuration such that the number of words $W$ is constant across configurations.
The above method enables the representation of motion patterns in query videos at different spatial locations and scales. When retrieving matches for query videos, we compute matches for all the 21 configurations and pool the results for further stages of annotation transfer and warping as described in Section \ref{sec:Warp}.  This enhanced flexibility leads to a reduction in size of the training video library required to represent arbitrary object motion in query videos. We then divide the documents of the query video into fragments as described in Section \ref{section:DocGen}.


\subsubsection{Library search and composition}
Consider a fragment of one of the query video documents:
\begin{equation}
f_q = {(w, \tau): w \in [0, W), \tau \in [0, T_f)}
\end{equation}

We wish to find a set of result fragments from our library, $\mathcal{F}_r$, which composed together approximate the query fragment:

\begin{equation}
\mathcal{F}_r = \argmax_{\mathcal{F}^{'}_{r}}\sum_{w}\sum_{\tau} \text{min} (R_{f_q}(w, \tau), R_{f_R}(w, \tau))
\end{equation}

\noindent where,

\begin{equation}
f_R = \bigcup_{f_r \in \mathcal{F}^{'}_{r}} f_r
\end{equation}

\begin{equation}
R_f(w, \tau) = \begin{cases}
  \frac{1}{|f|} , & \text{if } (w, \tau) \in f, \\
  0, & \text{otherwise}.
\end{cases}
\end{equation}

Here, $f_R$ is the union of all the selected result documents and $R_f(w, \tau)$ is a function that represents a set $f$ as a uniformly weighted, discrete probability distribution whose support is the $(w, \tau)$ pairs in $f$. As such, we are searching for the set of library fragments where  probability distribution for their union has a maximal histogram intersection with probability distribution for the query fragment. This can be rewritten as:

\begin{equation}
\mathcal{F}_r = \argmax_{\mathcal{F}^{'}_{r}} \frac{|f_q \cap f_R|}{\text{max} (|f_q|, |f_R|)}
\end{equation}

\begin{figure}
\centering
\includegraphics[width=\columnwidth]{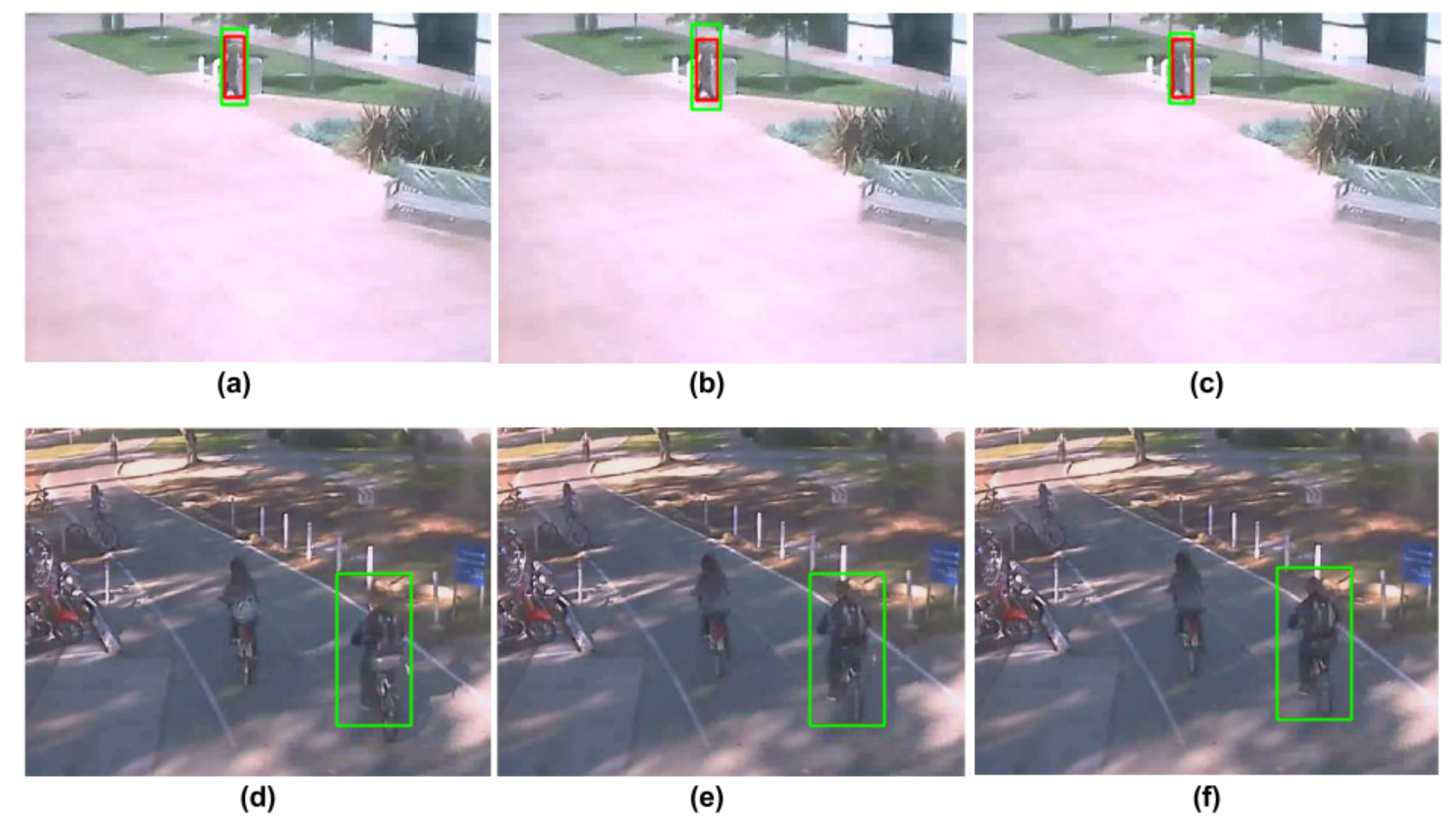}
\caption{An example of retrieval results for a query video sequence. Images (a)-(c) are frames from a query video and (d)-(f) represent frames from the top retrieved result amongst the library videos. The motion of the walking person in the query video in the up-left direction has been matched to the motion of the bicyclist. Note the difference in the spatial scales and locations of the objects in the query and result videos. Red boxes in (a)-(c) signify detected bounding boxes and green boxes signify ground-truth. Green boxes in (d)-(f) show the human annotated bounding boxes stored with the library videos. $\textit{Best viewed in color.}$}
\label{fig:exampleRetrieval}
\end{figure}

Choosing the library fragments  to include in the result set $\mathcal{F}_r$ is very similar to the maximum set coverage problem, which is NP-hard ~\cite{Hochbaum:1985:ASC:2455.214106}. We approach the selection of $\mathcal{F}_r$ using a greedy algorithm, which at each step adds a new fragment from the set of library fragments to the result set such that resulting histogram intersection is maximized. The retrieval algorithm is summarized in Algorithm \ref{algo:composition1}. In detail, a set of fragments from the library $X$, which share activations with the query fragment $f_q$ are retrieved using the Fragment reverse index $I_i$. We then find the fragments within $X$ which together compose $f_q$ in a greedy fashion. In the case where library videos are provided as queries, the algorithm will produce an exact match in the first iteration and generated tracks will be the same as ground truth. See Figure \ref{fig:exampleRetrieval} for an example of one of the library fragments retrieved for a query fragment.

The retrieval algorithm scales with multiple objects in the query video. Consider an example where we have two objects moving in a frame, one moves to the left and the other to the right.  Since the objects would occupy distinct spatial locations and would have different directions of motion, the activations get encoded in distinct locations of the corresponding document and consequently the fragments. 
This leads to two distinct motion patterns in the fragment. Each of the distinct patterns would result in retrieval results which compose these results independently. The design of the retrieval algorithm ensures that we get multiple composed fragments from the reference result with one corresponding to motion to the left and the other corresponding to motion to the right.

\begin{algorithm}
  \caption{Greedy Composition of Library Fragments: }\label{algo:composition1}
  $\textbf{Input:}$ \\
  \text{Query fragment $f_q$} \\
  \text{Fragment forward index} $I_f$ \\
  \text{Fragment inverse index} $I_i$ \\
  \text{Stopping criteria} $\rho$ \\ \\ 
  $\textbf{Output:}$ \\
  \text{Result fragment set} $\mathcal{F}_r$
  \begin{algorithmic}[1]
   
   \State $U \gets f_q$
   \State $U_0 \gets |U|$
   \State $\mathcal{F}_r \gets \left\{\right\}$
   \State $f_R \gets \left\{\right\}$
   \State $h \gets 0$
   
   \While{$|U| > \rho \ U_0$}
        \State $X \gets \bigcup_{(w, \tau) \in U} I_i[(w, \tau)]$
        \State $y \gets \{\}$
        \For{$x \in X$}
            
            \State $f_c \gets f_R \cup I_f[x]$
            \State $h \gets \frac{|f_q \cap f_c| }{\text{max}(|f_q|, |f_c|)}$
            \State $y \gets y \cup (h,x)$
            
        \EndFor
        \State $h_m, x_m \gets \text{max}(y)$
        
        \State $f_R \gets f_R \cup I_{f}[x_m]$
        \State $U \gets f_q \backslash f_R$
        \State $\mathcal{F}_r \gets \mathcal{F}_r \cup {x_m}$
        
   \EndWhile

  \end{algorithmic}
\end{algorithm}

\begin{figure}

\includegraphics[width=\columnwidth]{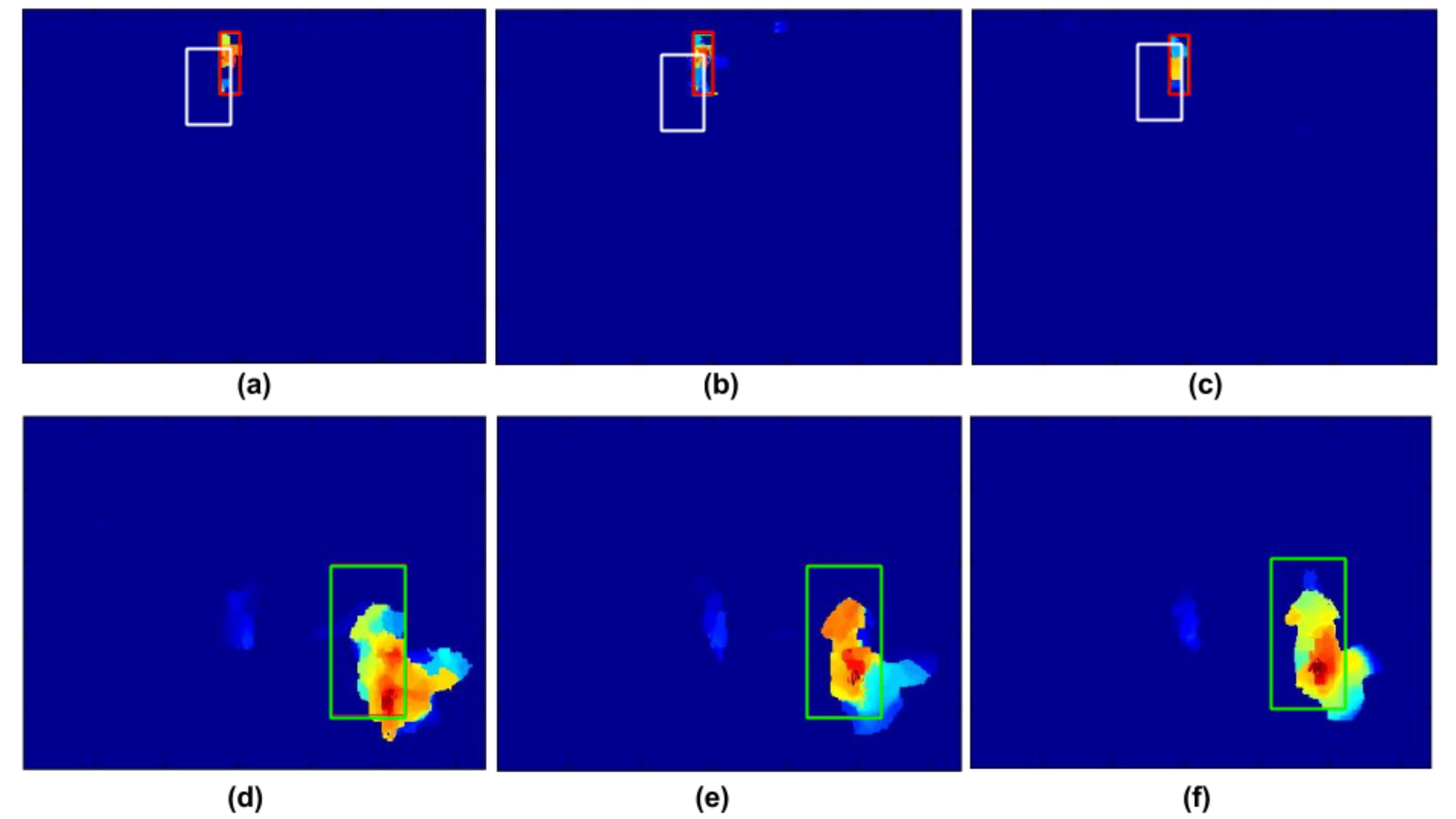}
\caption{Optical flow-fields for frames presented in Figure \ref{fig:exampleRetrieval} are shown in images (a)-(f). Query frame flow-fields (a)-(c) have the directly transferred bounding boxes highlighted in white  and results of the warping method (described in Section \ref{sec:Warp}) are drawn using red bounding boxes. Green boxes in (d)-(f) represent manually annotated bounding boxes from the retrieval results. $\textit{Best viewed in color.}$ }
\label{fig:retrievalExOF}
\end{figure}

\subsubsection{Annotation transfer and warping}\label{sec:Warp}

The previous step resulted in $\mathcal{F}_R$, the set of library result fragments that together best approximate the query fragment. Looking up each of these fragment names in $\textit{track inverse index}$ gives the set of unique track ids occurring in the result fragments, and looking up each of these up in the $\textit{track forward index}$ gives a set of bounding boxes to be transferred to the query video. Finally, we retrieve optical flow magnitude fields for the result fragments from the $\textit{flow fields}$ table. The indexes were previously defined in \ref{subsec:IndexDesc}. Each fragment corresponds to $T_f$ flow fields.

Figure \ref{fig:retrievalExOF} shows the flow fields and annotations retrieved for the example shown in Figure \ref{fig:exampleRetrieval}. Notice that while motion of the bicyclist in the results fragment and the pedestrian in the query fragment are similar, the objects are of different sizes and are in different locations in the image frame. We can not simply copy the bounding boxes from one to the other. Instead, the flow fields can be used to warp retrieved bounding boxes to better match the query.

It is not necessary to obtain a dense warping field from the result to the query; only the bounding box needs to be adjusted. The system seeks a bounding box on the query
flow field that is similar to the human-provided bounding box on the result flow field. This includes both the size and placement of the box, as well as the flow it contains. Bounding boxes are defined by their left, right, top and bottom edge positions and the system iteratively updates each edge of the query bounding box in turn to improve quality of the match. For a query bounding box $b_q$ and a result bounding box $b_r$ with left edge values $l_q$ and $l_r$, respectively, the update for $l_q$ is:
\begin{equation}
l_q = \argmax_{l^{'}_{q}}\Bigg[\sum_{i=0}^{N-1}\text{min}\bigg(H_{b_r}(i),H_{b_{q}^{'}}(i)\bigg)\Bigg]e^{\frac{-(l_r - l^{'}_{q})^{2}}{2\alpha}}
\label{eq:Warping}
\end{equation}

\noindent where $H_b$ is a $N$-bin normalized histogram of the flow magnitudes inside the bounding box $b$ and $\alpha$ is a penalty factor. As such, the update seeks a new query edge position which $(i)$ maximizes the histogram intersection between the histograms of the flows in $b_q$ and $b_r$, and $(ii)$ exacts a penalty for deviating too far from the result bounding box. The second part of the update criterion ensures that the query bounding box doesn't collapse onto a sub-region of the query frame's optical flow.  

The right, top and bottom edges proceed similarly. The warping scheme doesn't put a rigid constraint on the size of the final bounding box and allows adaptation to optical flow statistics of the local neighbourhood. Figure \ref{fig:retrievalExOF} shows an example result of warping bounding boxes. In our experiments, we randomly permute the order of the left, right top and bottom edges and obtain a batch of updates to eliminate bias that the order of edges might introduce. We have observed that the values of the edges converge reliably within 10 batches across multiple test matches. We have included a sensitivity analysis for $\alpha$ in Section $\ref{sec:alphaAnalysis}$.

Due to the overlapping nature of fragments, a frame can belong to multiple fragments. This leads to multiple bounding boxes being retrieved for a given motion pattern in a frame. To choose the best warped bounding box, we apply a non-maximal suppression rule to eliminate sub-optimal boxes. Bounding boxes are scored on the density of the optical flow being covered.

The chosen detection bounding boxes are associated together into object tracks by using the Hungarian Algorithm \cite{Munkres1, Munkres2} to solve an assignment problem where the association costs are modeled by a combination of geometric distance between bounding box centers and  color histogram distance. In detail, the association cost between bounding boxes $b^{n}_{i}, b^{n+1}_{j}$  in frames $n$ and $n+1$ are modelled as:
\begin{equation}
J_{ij}^{n, n+1} = d_{hist}(H^{\text{hsv}}_{b^{n}_{i}},  H^{\text{hsv}}_{b^{n+1}_{j}}) + \beta \norm{c_{b^{n}_i} - c_{b^{n+1}_j}}_2 
\end{equation}

\noindent where $H^{\text{hsv}}_b$ is the HSV color histogram of the image pixels lying within the bounding box $b$, $d_{hist}(.,.)$ is the histogram intersection distance, $\beta$ is a weight parameter, and $c_b$ is the center location of the bounding box $b$. The color histograms are constructed by jointly binning Hue and Saturation values. H and S channels are quantized into 10 and 5 equally spaced bins respectively. The parameter $\beta$ is fixed to 2.5 in our experiments, as due to poor image quality in our query videos, color information can be unreliable and provides only coarse discriminative information for association. 

Once tracks are generated from the above step, we perform post-processing in the form of a moving average filter with window width of $\pm$2 frames. We perform this step to improve temporal coherence of the generated bounding boxes. The averaging operation is carried out on center location and scale of the bounding boxes independently.

\section{Experiments}\label{section:Experiments}

\subsection{Datasets}
We have focused our experiments on surveillance videos. As the proposed approach is designed to be effective for low-quality, low-resolution videos, we have collected an appropriate dataset with 15 sequences. We call it the \textit{UCSB-Courtyard} dataset. These video clips have been recorded using Cisco WVC2300 wireless ip-network cameras overlooking a busy pedestrian crossing from five different view-points. Each sequence contains on an average 150 frames with pedestrians on a busy courtyard in an uncontrolled setting. The number of pedestrians vary from 1 to 4. The tracking targets undergo complex appearance changes due to shadows, occlusions and compression artifacts. The \textit{Browse2}, \textit{WalkByShop1front}, \textit{ShopAssistant1front}, \textit{TwoEnterShop2cor}, \textit{OneShopOneWait2cor} and \textit{OneLeaveShop1cor} sequences from CAVIAR \cite{CAVIAR} dataset are used for comparisons as well. These datasets are used to measure single object tracking performance. The proposed method is also capable of detecting multiple moving objects in a scene. To compare and benchmark with respect to other multiple object trackers, we have chosen the \textit{S2L2} sequence of PETS2009 \cite{PETS2009}. 

As described earlier, we have composed the library videos from a dataset which covers bike paths on a university campus. The scenes captured on this dataset are distinct from test datasets. We demonstrate that with a small library of videos, we can apply learnt motion patterns from one dataset onto an entirely different dataset. 

\subsection{Evaluation Metrics}

To perform quantitative comparison of object tracking, we use the standard metrics of Pascal Visual Object Challenge (VOC) detection score ~\cite{Everingham:2010:PVO:1747084.1747104} and Center Location Error (CLE) \cite{Wu_2013_CVPR}.

VOC score measures the quality of overlap between detected and ground-truth bounding boxes. For comparison of VOC scores of competing methods in a test dataset, we average the scores over frames in a sequence, and then over sequences to get the final score to generate a mean VOC score. 
CLE measures euclidean distance between the center of the detected bounding box and that of the ground-truth bounding box. CLE quantifies the localization ability of an object tracker. Similar to  mean VOC score, we calculate mean CLE score.

In addition to the above single object tracking metrics, we have also used CLEAR metrics ~\cite{bernardin2008evaluating} for comparison of algorithm performance with multiple object trackers. MOTP measures the ability to detect precise object locations whereas MOTA measures the capability of trackers to maintain consistent object configurations as  targets move around in the scene.

\begin{figure}

\includegraphics[width=\columnwidth]{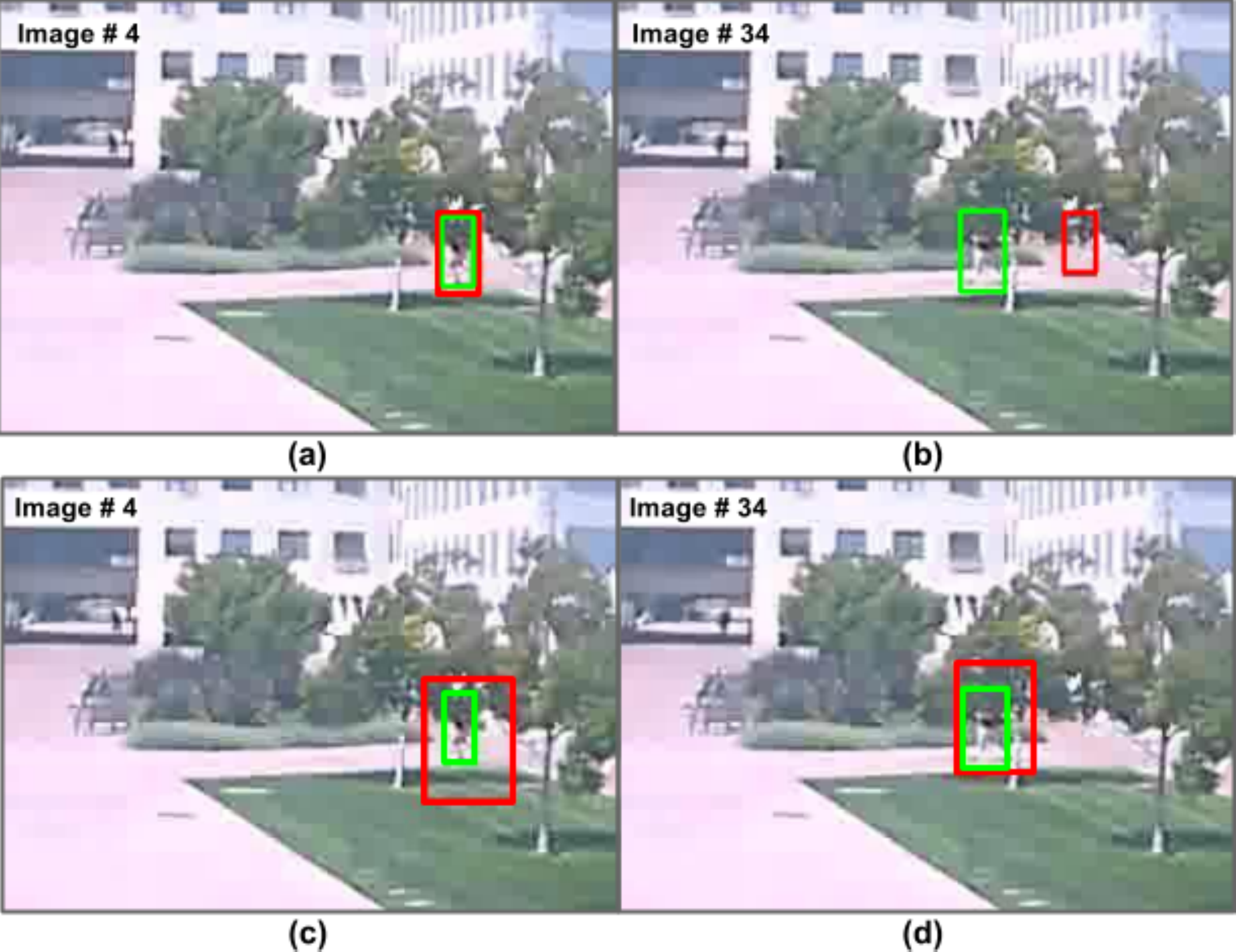}
\caption{(c)-(d) Results of ST on frames from the Courtyard dataset. ST has ignored scene clutter and continues to track the target across frames. (a)-(b) Results for VTD tracker are reproduced here from Figure \ref{fig:trackerFailure} for comparison.  Tracker results and ground truth boxes are marked in red and green respectively. $\textit{Best viewed in color.}$}
\label{fig:stSuccess}
\end{figure}

\subsection{Comparison with state-of-the-art}

In order to demonstrate advantages of the proposed approach over more conventional appearance-based approaches, we have chosen six state-of-the-art methods for comparison:

\begin{itemize}
\item \textit{Visual  Tracking  Decomposition} (VTD) \cite{DBLP:conf/cvpr/KwonL10}: This method combines multiple appearance-based observation model and  motion model trackers using Sparse Principle Component Analysis
and an Interactive Markov Chain Monte Carlo framework. An initial bounding box of the target is required for tracking.
\item \textit{Struck  Tracking} \cite{struckTracker}: This adaptive method formulates the problem of choosing good training examples for online training of target appearance as a structured SVM. An initialization of the target position is required for tracking.

\item \textit{Adaptive Color Tracking} (ACT) \cite{diva2:711538}: This real-time tracking method incorporates sophisticated color features to provide invariant representation in the illumination space. An initial bounding box of the target is required for tracking.

\item \textit{Initialization-Insensitive Tracking} (IIT) \cite{InitInsensitive}: This approach utilizes motion saliency of local features to accurately track objects in an adaptive manner with inaccurate initializations. Target position initialization is required here as well.

\item \textit{Consensus-based Tracking and Matching of Keypoints for Object Tracking } (CMT) \cite{Nebehay2015CVPR}: This method tracks feature points across frames to estimate target location in current frame. Target position initialization is a requirement.

\item \textit{Background Subtraction-based Tracking} (BGS) \cite{Kaewtrakulpong01animproved}: This method segments out moving objects in a scene from the background and applies a Kalman Filter over bounding box estimates.

\end{itemize}

The results for the competing methods have been generated using codes provided by respective authors, with parameters set to the default values suggested by provided documentation.

\begin{figure*}
\begin{subfigure}{.5\textwidth}
  \centering
  \includegraphics[width=\linewidth]{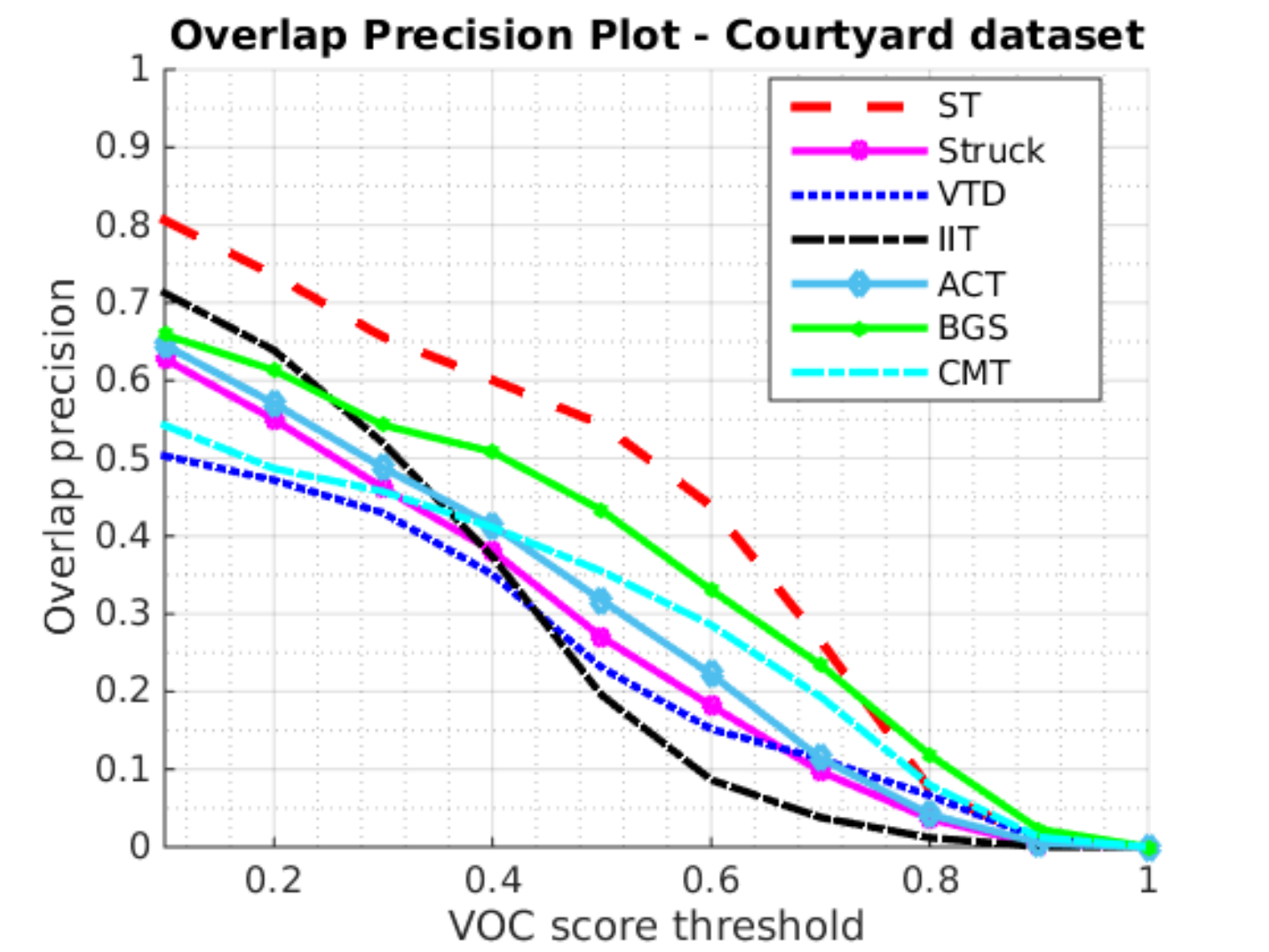}
  \caption{}
  \label{fig:overlapCourtyard}
\end{subfigure}%
\begin{subfigure}{.5\textwidth}
  \centering
  \includegraphics[width=\linewidth]{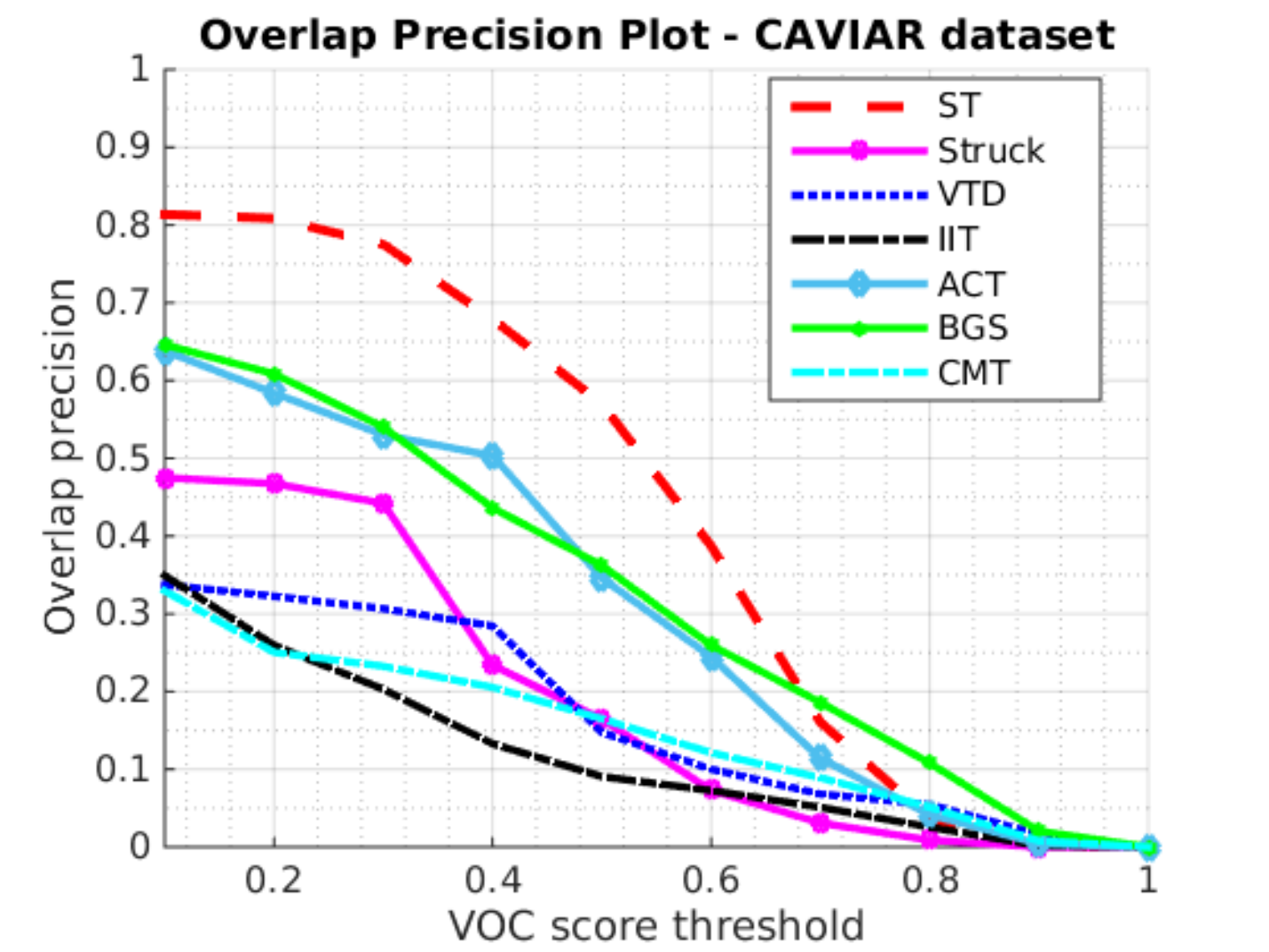}
  \caption{}
  \label{fig:overlapCAVIAR1}
\end{subfigure}
\begin{subfigure}{.5\textwidth}
  \centering
  \includegraphics[width=\linewidth]{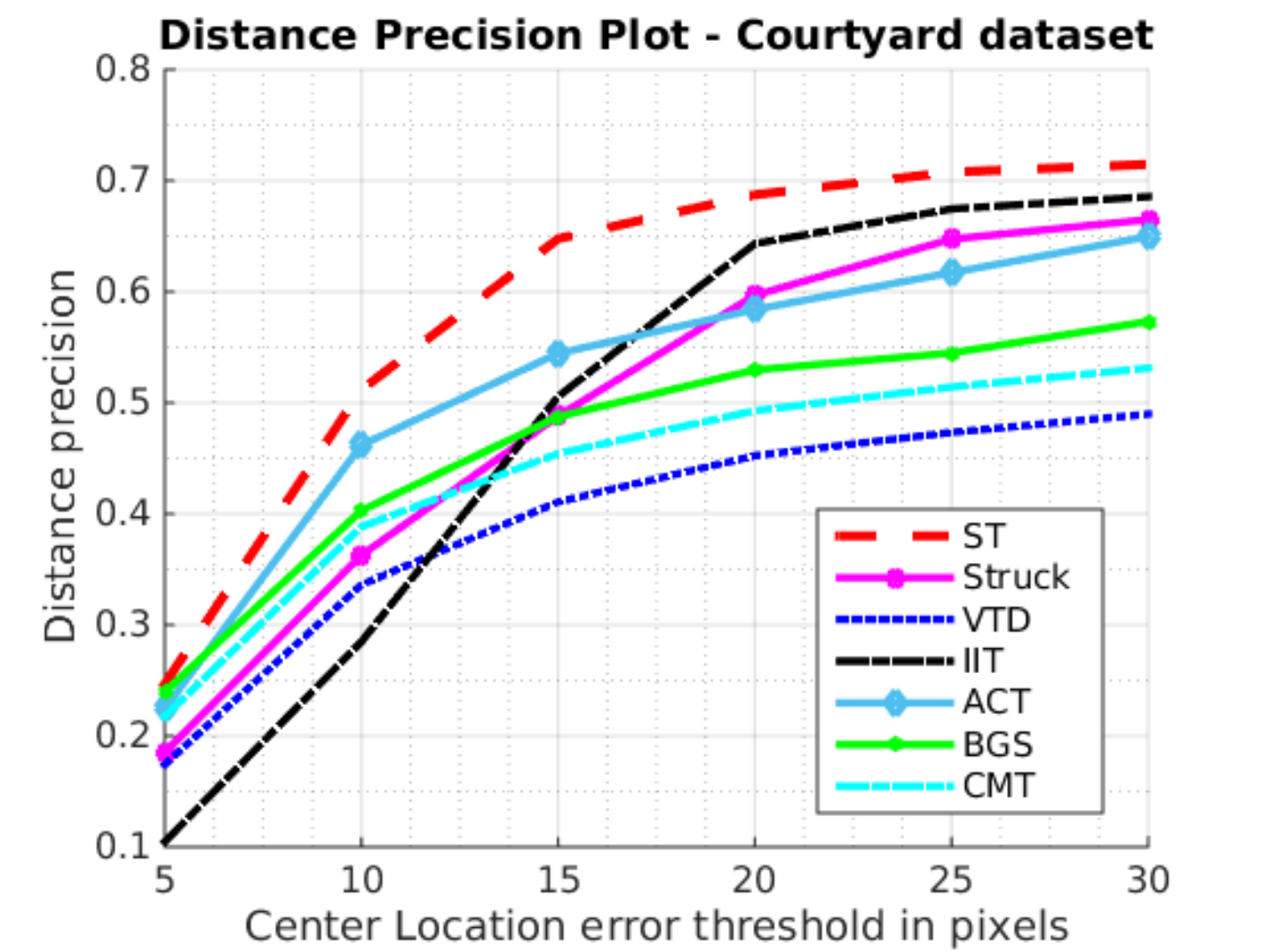}
  \caption{}
  \label{fig:overlapCAVIAR2}
\end{subfigure}
\begin{subfigure}{.5\textwidth}
  \centering
  \includegraphics[width=\linewidth]{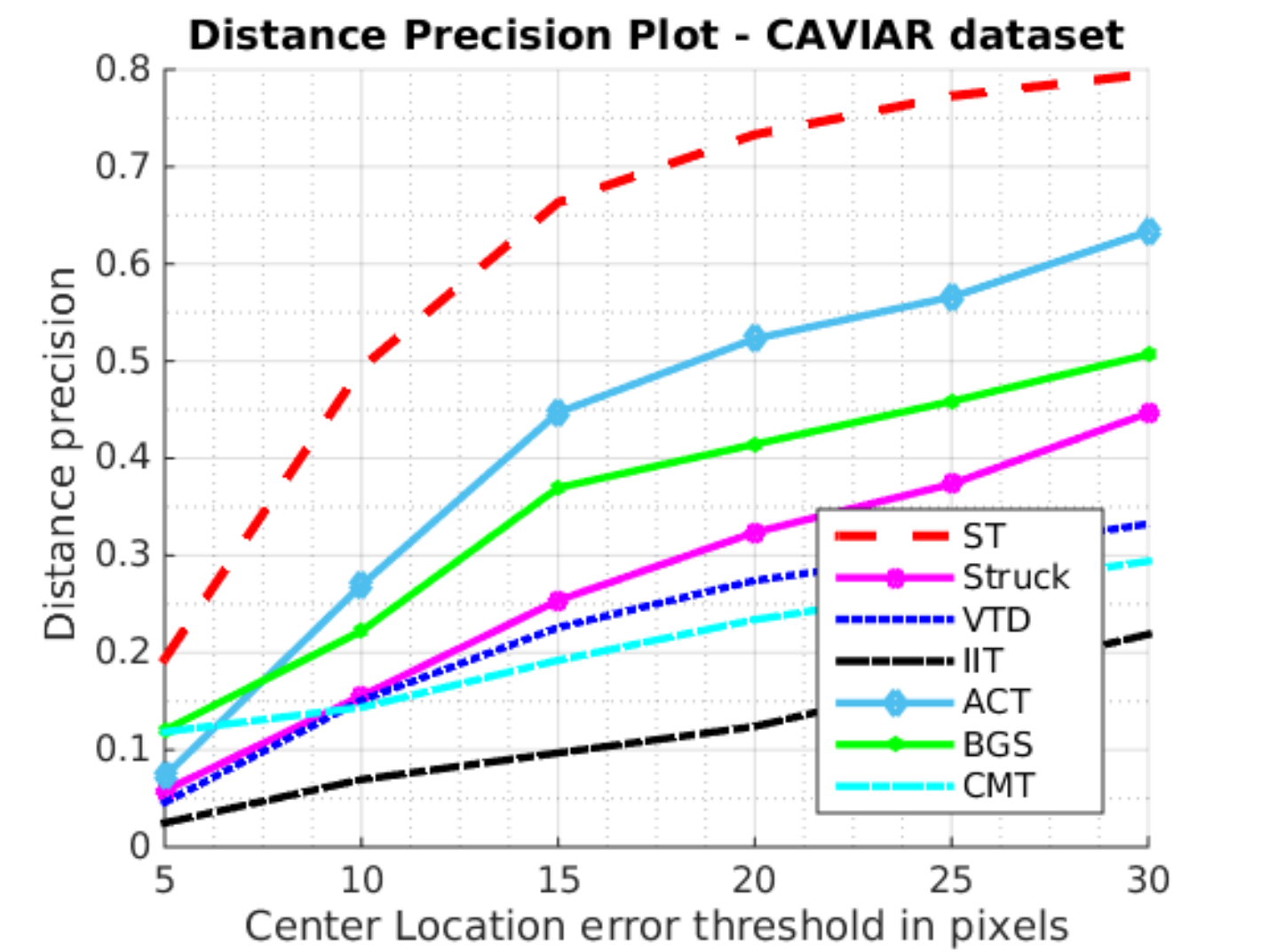}
  \caption{}
  \label{fig:overlapCAVIAR3}
\end{subfigure}
\caption{Comparative distance and overlap precision performance plots for Search Tracker (ST) and competing algorithms on the Courtyard and CAVIAR datasets.}
\label{fig:precisionPlots}
\end{figure*}

Tables \ref{table:meanVOC} and \ref{table:meanCLE} show comparison of mean VOC and mean CLE scores across different datasets between the proposed method and competing methods. Tables \ref{table:overlapPrecision} and \ref{table:distancePrecision} report comparative results on mean overlap precision and mean distance precision across datasets and methods. The distance and overlap thresholds are set to 20 pixels and 0.5 respectively. Figure \ref{fig:precisionPlots} presents distance and overlap precision scores for different values of VOC score and CLE thresholds. ST consistently outperforms all other competing algorithms by a wide margin.

\begin{figure*}
\begin{center}
\includegraphics[width=2\columnwidth]{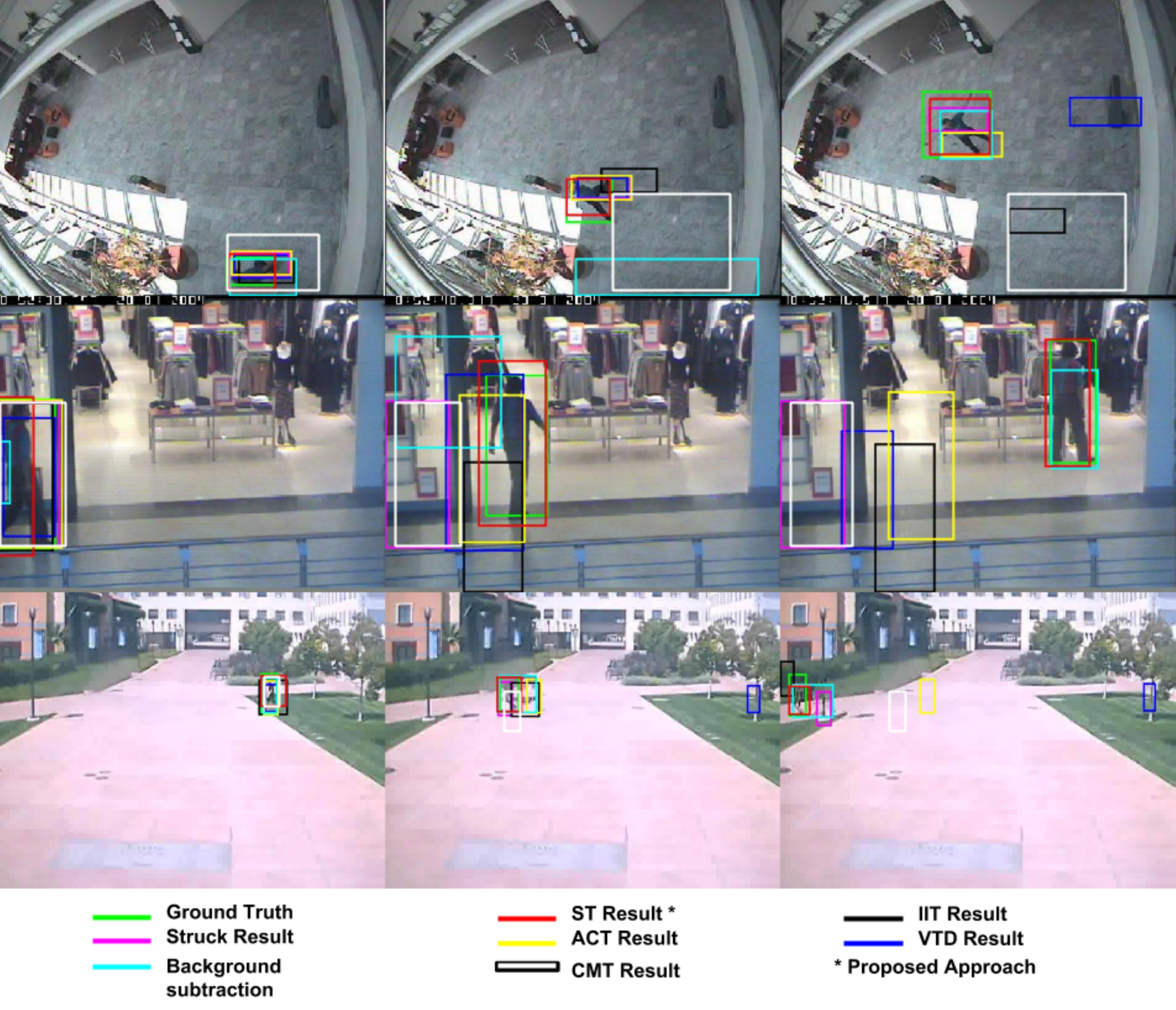}
\end{center}
   \caption{Comparison of our approach with state-of-the-art trackers on CAVIAR and Courtyard datasets. The first (top) row shows images where the target undergoes illumination and shape variations. In the second row, the target passes through a cluttered scene. The target in the third row undergoes compression artifacts and an occlusion. The proposed tracker is able to track the targets and it adapts bounding box scale to the target size, whereas the competing trackers get distracted by scene clutter, have fixed bounding box scales, and fail when the target appearance changes or undergoes occlusions. $\textit{Best viewed in color.}$}
\label{fig:comparison}
\end{figure*}

As we can see, ST is competitive with respect to the  appearance-based methods. It is important to note that we do not depend on manually provided initial bounding boxes or object detectors for the training videos. This gives us a strong advantage when the manual initialization or good object detectors are not available especially in test datasets suffering from poor image-quality. ST outperforms competing methods by a large margin in the Courtyard dataset.  We are able to get this performance from ST without any manual initialization. CAVIAR has indoor sequences set in a shopping mall with comparatively low image quality and  more scene clutter. Therefore, leveraging motion patterns helps us outperform all the other algorithms on CAVIAR. Example result frames are presented in Figure \ref{fig:comparison}. These frames illustrate resilience of our algorithm to scene clutter, illumination changes and occlusions. In addition, aforementioned image-quality issues often cause BGS methods to  fail on both the CAVIAR and the Courtyard sequences. To contrast against the usage of optical flow and feature tracking methods, we have provided comparisons with the CMT tracker. Due to poor image quality of test videos, consistent tracking of object feature points across multiple points is a difficult problem and hence leads to comparatively weaker tracker performance. Since ST utilizes aggregated optical flow information across multiple frames at the same time, the tracker is robust to such conditions.

In Figure \ref{fig:trackerFailure}, we presented a case where scene clutter and compression artifacts can cause appearance-based trackers to fail. ST can overcome such issues since quantized long-term motion patterns are robust to the presence of scene clutter and occlusions. Results for ST on the same sequence are shown in Figure \ref{fig:stSuccess}.  Search Tracker is also robust  to abrupt appearance changes due to shadows, compression artifacts, and  changing illumination because of the relative invariance of long-term motion patterns, whereas appearance-based trackers frequently fail in such sequences. These issues are very important to address as they are commonplace in real world scenarios.

In order to compare the performance with respect to tracking multiple objects, we provide the bounding boxes generated by the Search Tracker on the PETS 2009 S2L2 sequence to $\cite{milan2013detection}$ which combines detections into object tracks using an energy minimization framework. We compute the MOTA and MOTP scores generated for these tracks and compare them with the state-of-the-art in Table $\ref{table:MOTAP}$. Our method is comparable in performance with other multi-object trackers. A point to note is that the competing methods use external sources for object bounding boxes.


\begin{table}[h]

\begin{tabular}{|l|l|l|l|}
\hline
\textbf{Algorithm}      & \textbf{Courtyard} & \textbf{CAVIAR} \\ \hline
$\text{Struck Tracker}^{\mathbf{\ast}} \cite{struckTracker}$ & 0.2952          & 0.1338    \\ \hline
$\text{VTD}^{\mathbf{\ast}} \cite{DBLP:conf/cvpr/KwonL10}$            & 0.2593          & 0.0858    \\ \hline
$\text{ACT}^{\mathbf{\ast}} \cite{diva2:711538}$            & 0.3163          & 0.3291    \\ \hline
$\text{IIT}^{\mathbf{\ast}} \cite{InitInsensitive}$            & 0.2949          & 0.0743    \\ \hline
$\text{BGS} \cite{Kaewtrakulpong01animproved}$                            & 0.3803          & 0.4160    \\ \hline
$\text{CMT}^{\mathbf{\ast}} \cite{Nebehay2015CVPR}$                            & 0.3114          & 0.1661    \\ \hline

$\textbf{Search Tracker}$   & \textbf{0.4539}             & \textbf{0.6075}    \\ \hline
\end{tabular}
\caption {Comparative table of mean VOC scores for datasets across tracking methods. A higher value reflects superior tracking results. Algorithms marked with $\mathbf{\ast}$ require manual initialization. }
\label{table:meanVOC}
\end{table}

\begin{table}[h]

\begin{tabular}{|l|l|l|l|}
\hline
\textbf{Algorithm}      & \textbf{Courtyard} & \textbf{CAVIAR} \\ \hline
$\text{Struck Tracker}^{\mathbf{\ast}} \cite{struckTracker}$ & 37.41          & 75.38    \\ \hline
$\text{VTD}^{\mathbf{\ast}} \cite{DBLP:conf/cvpr/KwonL10}$            & 73.57          & 89.60    \\ \hline
$\text{ACT}^{\mathbf{\ast}} \cite{diva2:711538}$            & 28.54           & 42.66    \\ \hline
$\text{IIT}^{\mathbf{\ast}} \cite{InitInsensitive}$             & 24.32          & 82.52    \\ \hline
$\text{BGS}\cite{Kaewtrakulpong01animproved} $                            & 49.13          & 38.39    \\ \hline
$\text{CMT}^{\mathbf{\ast}} \cite{Nebehay2015CVPR}$             & 59.54          & 81.66    \\ \hline
$\textbf{Search Tracker}$   & \textbf{21.06}             & \textbf{27.46}    \\ \hline
\end{tabular}

\caption {Comparative table of mean CLE scores in pixels for datasets across tracking methods. A lower value reflects superior tracking results. Algorithms marked with $\mathbf{\ast}$ require manual initialization.}
\label{table:meanCLE}
\end{table}

\begin{table}[h]

\begin{tabular}{|l|l|l|l|}
\hline
\textbf{Algorithm}      & \textbf{Courtyard} & \textbf{CAVIAR} \\ \hline
$\text{Struck Tracker}^{\mathbf{\ast}} \cite{struckTracker}$ & 0.2698          & 0.1645    \\ \hline
$\text{VTD}^{\mathbf{\ast}} \cite{DBLP:conf/cvpr/KwonL10}$            & 0.2306          & 0.1473    \\ \hline
$\text{ACT}^{\mathbf{\ast}} \cite{diva2:711538}$            & 0.3172          & 0.3459    \\ \hline
$\text{IIT}^{\mathbf{\ast}} \cite{InitInsensitive}$            & 0.1951          & 0.0906    \\ \hline
$\text{BGS}\cite{Kaewtrakulpong01animproved} $                            & 0.4330          & 0.3621    \\ \hline
$\text{CMT}^{\mathbf{\ast}}\cite{Nebehay2015CVPR} $                            & 0.3124          & 0.1649    \\ \hline

$\textbf{Search Tracker}$   & \textbf{0.5433}             & \textbf{0.5700}    \\ \hline
\end{tabular}

\caption {Comparative table of mean overlap precision for datasets across tracking methods. The overlap threshold is set to 0.5. A higher value reflects superior tracking results. Algorithms marked with $\mathbf{\ast}$ require manual initialization.}
\label{table:overlapPrecision}
\end{table}

\begin{table}[h]

\begin{tabular}{|l|l|l|l|}
\hline
\textbf{Algorithm}      & \textbf{Courtyard} & \textbf{CAVIAR} \\ \hline
$\text{Struck Tracker}^{\mathbf{\ast}} \cite{struckTracker}$ & 0.5968          & 0.3238    \\ \hline
$\text{VTD}^{\mathbf{\ast}} \cite{DBLP:conf/cvpr/KwonL10}$           & 0.4523          & 0.2743    \\ \hline
$\text{ACT}^{\mathbf{\ast}} \cite{diva2:711538}$            & 0.5839          & 0.5230    \\ \hline
$\text{IIT}^{\mathbf{\ast}} \cite{InitInsensitive}$           & 0.6432          & 0.1240    \\ \hline
$\text{BGS}\cite{Kaewtrakulpong01animproved} $                            & 0.5295          & 0.4145    \\ \hline
$\text{CMT}^{\mathbf{\ast}}\cite{Nebehay2015CVPR} $                            & 0.5197          & 0.2340    \\ \hline
$\textbf{Search Tracker}$   & \textbf{0.6873}             & \textbf{0.7331}    \\ \hline
\end{tabular}

\caption {Comparative table of mean distance precision for datasets across tracking methods. The distance threshold is set to 20 pixels. A higher value reflects superior tracking results. Algorithms marked with $\mathbf{\ast}$ require manual initialization.}
\label{table:distancePrecision}
\end{table}




\begin{table}[h]
\begin{tabular}{|l|l|l|}
\hline
Algorithm      & \textbf{MOTA}         & \textbf{MOTP}          \\ \hline
Milan et al.$^{\mathbf{\ast}}\cite{milan2014continuous}$    & 56.9          & \textbf{59.4} \\ \hline
$\text{JPDA}_{100}$ $^{\mathbf{\ast}}\cite{Rezatofighi:2015:ICCV}$           & \textbf{59.3} & 58.27         \\ \hline
Search Tracker & 46.4          & 55.8          \\ \hline
\end{tabular}
\caption {Comparative table of CLEAR MOT scores for PETS-2009 S2L2 sequence across tracking methods. A higher value reflects superior tracking results. Note that both metrics are normalized such that  scores of 100 for both MOTA and MOTP correspond to perfect alignment of tracker generated bounding boxes to ground truth and no identity switches amongst tracks. Algorithms marked with $\mathbf{\ast}$ require external object detectors.}
\label{table:MOTAP}
\end{table}

\begin{figure}
\begin{subfigure}{.25\textwidth}
  \centering
  \includegraphics[width=\linewidth]{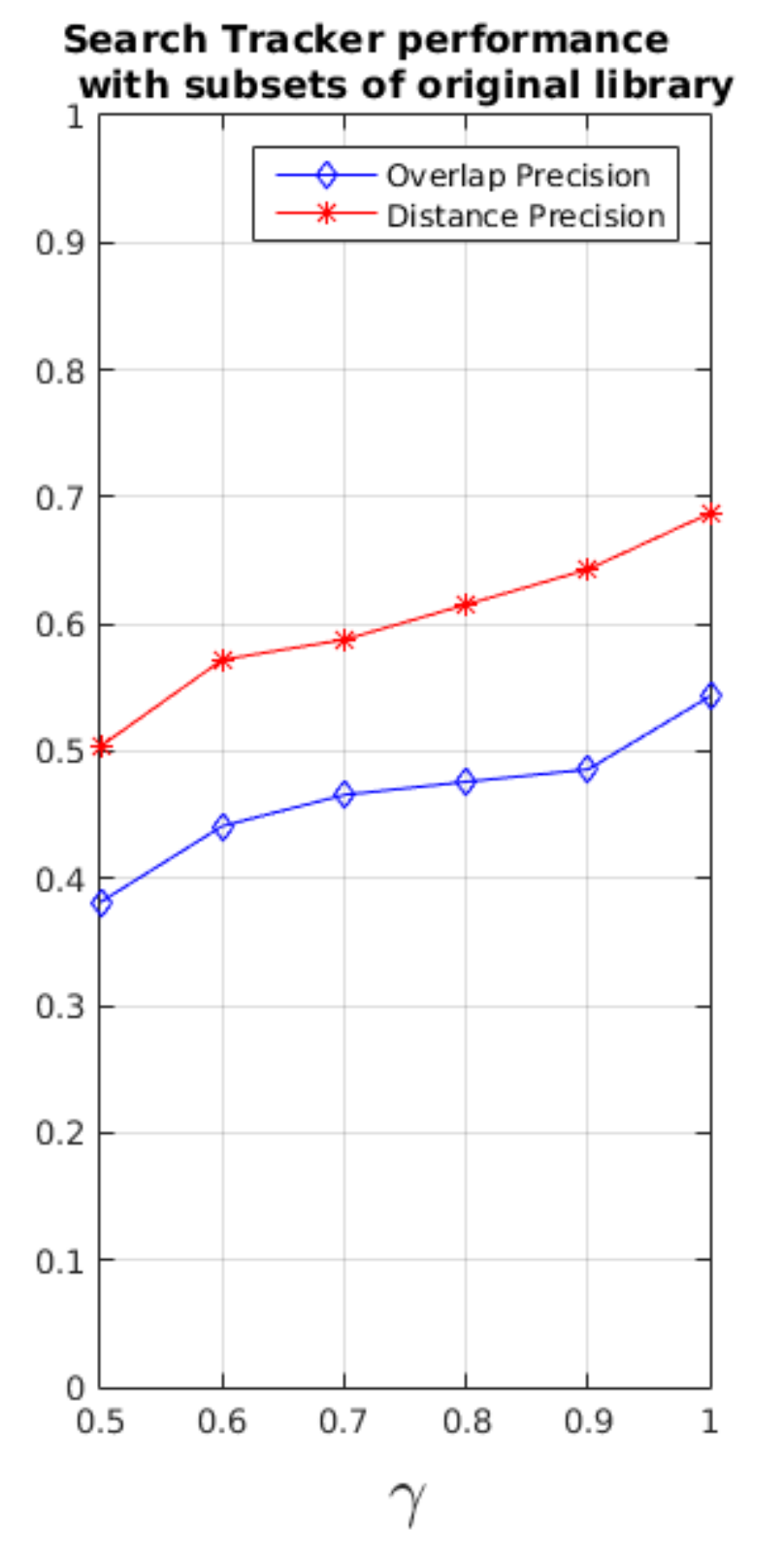}
  \caption{\small{}}
  \label{fig:sublib}
\end{subfigure}%
\begin{subfigure}{.25\textwidth}
  \centering
  \includegraphics[width=\linewidth]{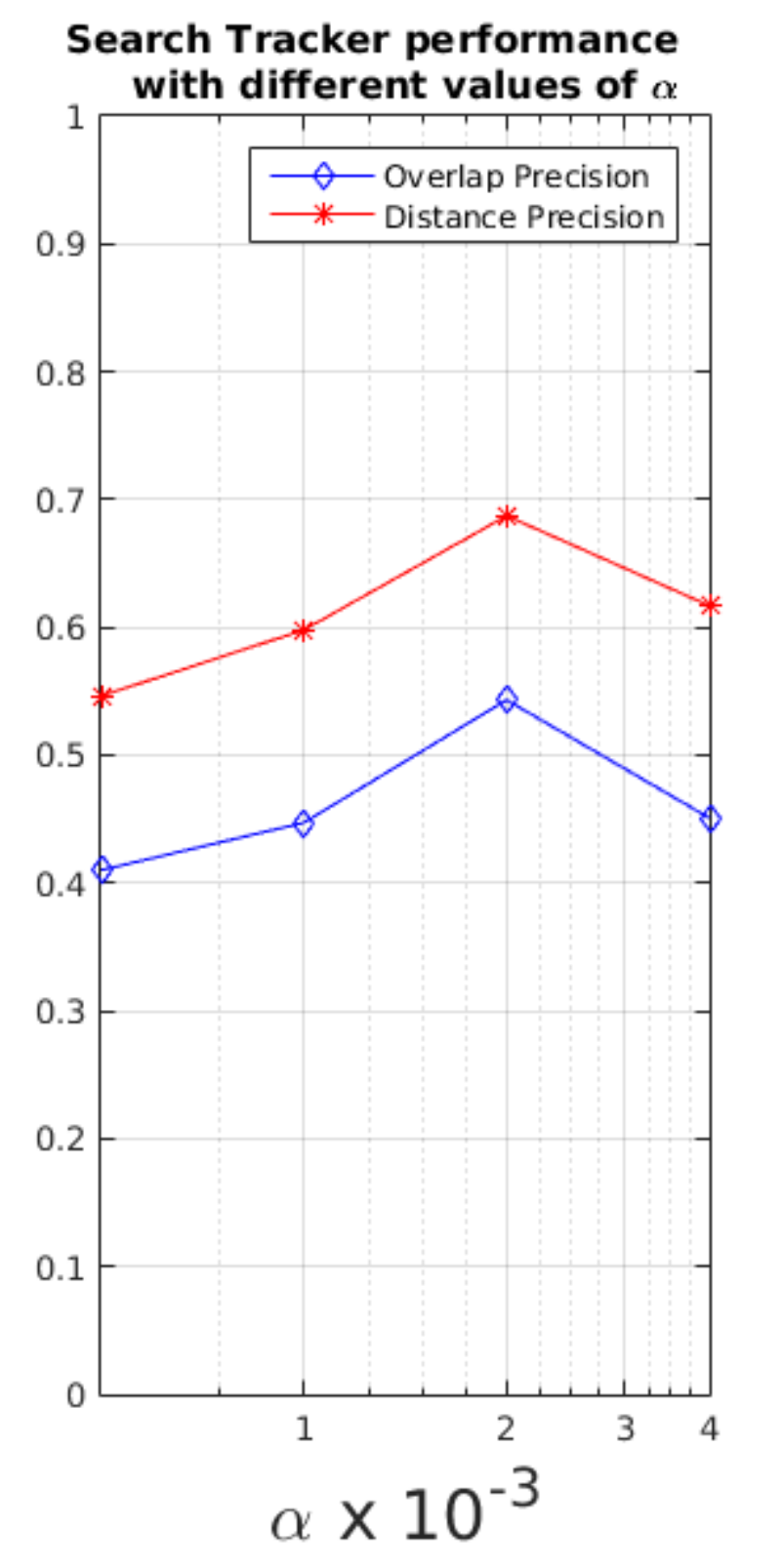}
  \caption{}
  \label{fig:alphaParamSweep}
\end{subfigure}
\caption{(a) Overlap precision and Distance precision values for the proposed method on the Courtyard dataset with $\gamma$ fraction of the entire video library used for the retrieval database. (b) Overlap precision and Distance precision values for the proposed method on the Courtyard dataset for different values of the parameter $\alpha$. The VOC score threshold and CLE thresholds were set as 0.5 and 20 pixels respectively for both of the plots and Courtyard was used as test dataset to generate both of the plots.}
\end{figure}

\subsection{Performance analysis with varying library sizes}
We investigate the effect of different library sizes on the proposed method's tracking performance. We chose randomly $\gamma = \{ 0.5, 0.6, 0.7, 0.8, 0.9, 1\}$ fraction of the library videos and generate sub-libraries. We then run the search and retrieval algorithm with one of these sub-libraries at a time, and plot the Overlap precision and the Distance precision scores on the Courtyard dataset for the different values of $\gamma$ in Figure $\ref{fig:sublib}$. As can be seen from the plots, ST's performance scales with size of the associated annotated video library. Since we apply data augmentation techniques in the form of vertical and horizontal flipping of library videos and also generate multi-scale query video representations, the proposed method's performance does not reduce by a large margin due to reduction in library sizes.

\subsection{Analysis on Annotation Warping} $\label{sec:alphaAnalysis}$
In the annotation warping stage, we control the flexibility that a transferred bounding box has in fitting  optical flow characteristics of the query video frame, through the penalty term $\alpha$ from equation $\ref{eq:Warping}$. We found the optimal value of $\alpha$ to be 2000 for our experiments.   To investigate the sensitivity of the proposed method for different values of $\alpha$, we execute the proposed tracker on the Courtyard dataset and measure the Overlap precision and  Distance precision at VOC score thresholds of 0.5 and 20 pixels respectively. The tracking performance of ST is shown in Figure $\ref{fig:alphaParamSweep}$. Low values of $\alpha$ restricts flexibility of the transferred bounding box to adapt the test sequence's optical flow characteristics, while higher values can lead to bounding boxes collapsing onto regions of high optical flow magnitude.

\subsection{Computational cost}
Our experiments were carried out on a single-core 3.5 Ghz workstation using MATLAB. The query stage and the bounding box composition steps take between 4 to 25 seconds for each frame, depending on the number of moving objects in the scene. The computational cost of ST is distributed amongst the query multi-scale fragment computation stage, the library search and composition stage and the annotation transfer and warping stage. The time required per frame for fragment generation is 53 msec, the library search stage needs 3.7 sec and the annotation transfer stage requires 9.3 secs on an average for the Courtyard dataset. 

Cost of fragment generation is independent of the content in query videos. Annotation transfer and warping requires the largest amount of computation amongst all the stages. Since a frame can be a member of multiple query fragments, the large number of matched annotations and the accompanying warping procedure adds to the computational cost.
Annotation warping  can be made faster by a parallelized implementation for warping of retrieved candidate bounding boxes. optical flow method provided by \cite{SecretsofOF} provided the most accurate results, but the method is computationally expensive and this adds to the cost of ST. 

\section{Discussion} \label{section:discConc}

There are a few limitations to the proposed method. ST is designed to work with stationary cameras and will not be directly applicable to data from PTZ and mobile-device cameras. There may be cases where the motion present in the test video cannot be modeled by the training library database, which can be overcome by adding more video clips to the library. Diversity can also be induced by generating translated and rotated versions of pre-existing library videos. Also, with state-of-the-art trackers becoming more efficient and robust, we could combine  automated tracker outputs instead of depending on human-generated annotations to create cheaper, large-scale video libraries and consequently lead to improved object tracking. We also expect that this method of directly transferring knowledge available on one annotated dataset to a different dataset to be applicable to other problems like action recognition, activity analysis and others task that can be analysed through motion patterns.

ST also has limitations with respect to modelling target motion in crowded sequences. In sequences where a large number of targets occlude each other, the optical flow signatures are not discriminative enough to find a good match from the library dataset. In some cases, very small objects in scenes do not generate strong optical flow fields and hence encoding of motion becomes challenging.  ST is best suited for tracking fewer number of objects in cluttered and challenging scenarios.

\section{Future Work and Conclusions}\label{section:futWork}

In the proposed method, the transferred annotations are warped on each frame from the query video. The warping algorithm could be made more robust and efficient by considering optical flow characteristics of adjacent frames, resulting in smoother tracks.

The paradigm of learning motion patterns and behaviours from an annotated library of past videos can be extended to several novel surveillance scenarios. Consider a surveillance network where we have annotations for videos from a subset of the connected cameras. With the remaining cameras or in the event of adding a new camera, we could directly start leveraging the past motion pattern knowledge mined from the annotated dataset.  ST could also be used in an active learning framework where imperfect appearance-based trackers and detectors are used as `teacher' algorithms to create a seed library. ST as the `student' algorithm tracks objects in conditions which are difficult for appearance-based trackers using the library. The library continuously expands, both from the past output of ST and appearance based trackers, which would lead to improvement of ST performance. The basic idea of similarity search of motion patterns could be explored for applications in action recognition, object retrieval and object re-identification from videos.

We have presented a novel approach to tracking that uses human annotations to directly drive an automated tracking system. We generate documents from videos which represent motion patterns. These documents are used to retrieve videos with similar motion characteristics and associated annotations are transferred and warped to the query video. This system avoids the requirement of object detectors and outperforms state-of-the-art appearance-based trackers on in-the-wild surveillance datasets, which has been demonstrated in the experiments.

\section{Acknowledgements}

This research is partially supported by the following three grants: (a) by the US Office of Naval Research N00014-12-1-0503, and (b) by the Institute for Collaborative Biotechnologies through grant W911NF-09-0001 from the U.S. Army Research Office,   
(c) by the Army Research Laboratory under Cooperative Agreement Number W911NF-09-2-0053 (the ARL Network Science CTA). The views and conclusions contained in this document are those of the authors and should not be interpreted as representing the official policies, either expressed or implied, of the Army Research Laboratory or the U.S. Government. The U.S. Government is authorized to reproduce and distribute reprints for Government purposes notwithstanding any copyright notation here on.



%





\ifCLASSOPTIONcaptionsoff
  \newpage
\fi

\newpage
\clearpage

\end{document}